\documentclass[pmlr]{jmlr}

\RequirePackage{graphicx}
\usepackage{booktabs}
\usepackage{longtable}
\usepackage{float}
\usepackage{xcolor}
\usepackage{wrapfig} 

\newif\iftrackchanges
\trackchangesfalse
\iftrackchanges
  \newcommand{\revised}[1]{{\color{blue}#1}}
  \newcommand{\reviewnote}[1]{{\color{blue}\footnote{\textbf{[Rev]:} #1}}}
\else
  \newcommand{\revised}[1]{#1}
  \newcommand{\reviewnote}[1]{}
\fi

\makeatletter
\def\set@curr@file#1{\def\@curr@file{#1}}
\makeatother
\usepackage[load-configurations=version-1]{siunitx}

\theorembodyfont{\upshape}
\theoremheaderfont{\scshape}
\theorempostheader{:}
\theoremsep{\newline}

\jmlrproceedings{PMLR}{Proceedings of Machine Learning Research}
\jmlrvolume{340}
\jmlryear{2026}
\jmlrworkshop{Machine Learning for Healthcare}

\title[WILLIE: Wound Imaging, Localization, and Integrated Evaluation]{WILLIE: A Unified Framework and Benchmark \\for Wound Classification, Segmentation, and Localization}

\author{\Name{Gopi Trinadh Maddikunta}
       \Email{gmaddiku@cougarnet.uh.edu}\\ 
       \addr Department of Computer Science\\
       University of Houston\\
       Houston, Texas, USA 
       \AND
       \Name{Shannan Hamlin}
       \Email{SHamlin@houstonmethodist.org}\\
       \addr Houston Methodist Academic Institute\\
       Houston Methodist Hospital System\\
       Houston, Texas, USA
       \AND
       \Name{Hsin-Mei Chen}
       \Email{hchen5@houstonmethodist.org}\\
       \addr Houston Methodist Academic Institute\\
       Houston Methodist Hospital System\\
       Houston, Texas, USA
       \AND
       \Name{Kimaya Barnes}
       \Email{ksbarnes@houstonmethodist.org}\\
       \addr Houston Methodist Academic Institute\\
       Houston Methodist Hospital System\\
       Houston, Texas, USA
       \AND
       \Name{Peizhu Qian}
       \Email{pqian@uh.edu}\\ 
       \addr Department of Computer Science\\
       University of Houston\\
       Houston, Texas, USA
       } 

\newcommand\blfootnote[1]{%
  \begingroup
  \renewcommand\thefootnote{}\footnote{#1}%
  \addtocounter{footnote}{-1}%
  \endgroup
}

\begin{document}

\maketitle
\blfootnote{This is an extended version of the paper published in the
Proceedings of the 11th Machine Learning for Healthcare Conference
(MLHC 2026), PMLR 340:1243--1263, 2026.
\url{https://proceedings.mlr.press/v340/maddikunta26a.html}}

\begin{figure}[h]
\vspace{-4em}
  \centering
  \includegraphics[width=0.85\textwidth, height=5.5cm]{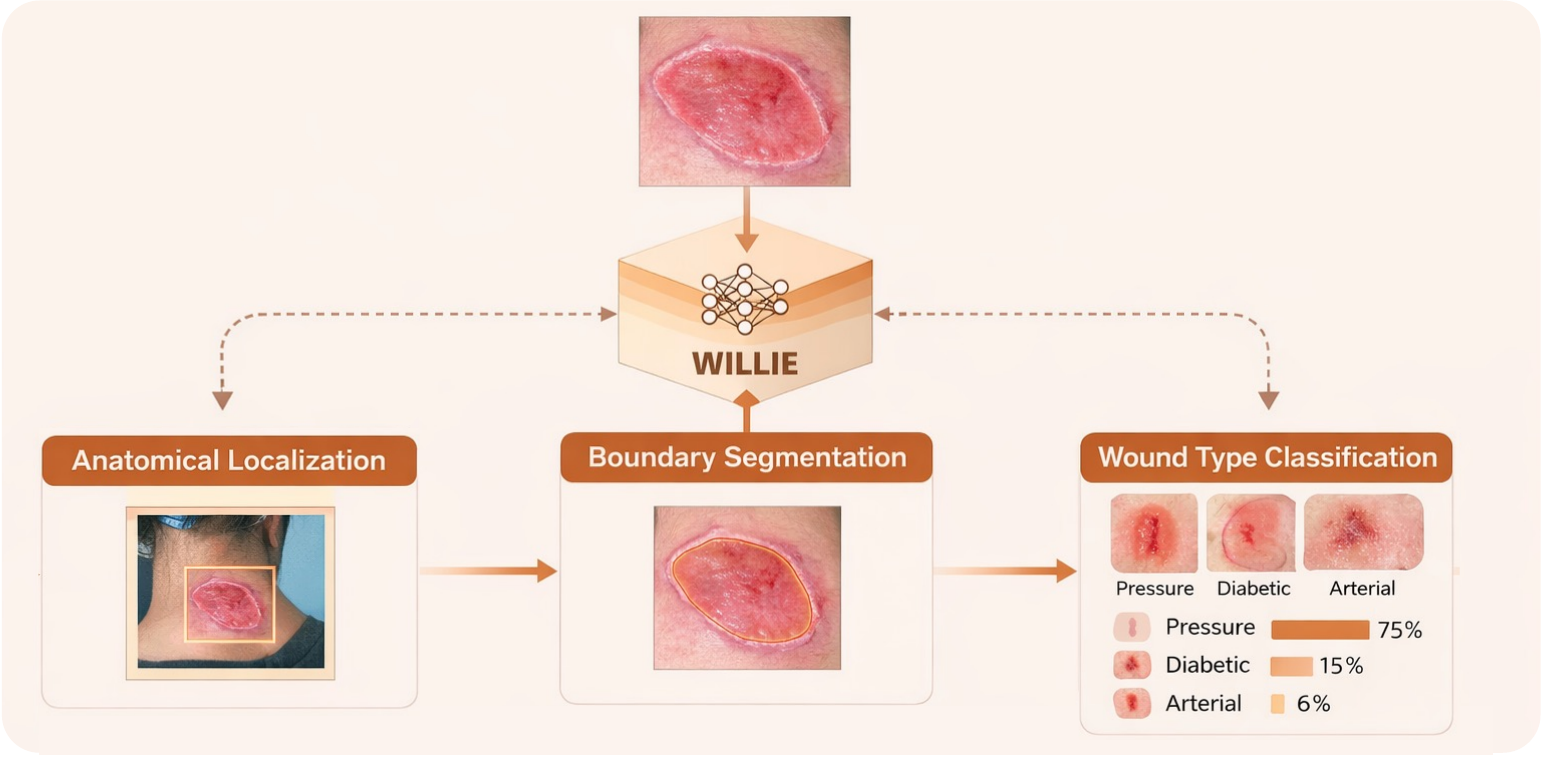}
  \vspace{-1em}
  \caption{Clinical wound assessment involves three coupled image analysis tasks: anatomical localization, boundary segmentation, and wound type classification. \texttt{WILLIE} provides a unified framework for producing these outputs with a single pipeline.}
  \label{fig:teaser}
\vspace{-1em}
\end{figure}

\begin{abstract}
Chronic wound management affects over 8.2 million patients in the United States and imposes substantial clinical and economic burden. 
Clinical wound assessment commonly involves three coupled tasks: identifying wound type, delineating wound boundaries, and localizing the wound region for measurement and monitoring. 
Despite this clinical coupling, existing machine learning approaches typically address wound classification, segmentation, and localization using separate models. 
We present \textbf{WILLIE}, a unified framework and benchmark for wound classification, segmentation, and localization that enables systematic evaluation of multi-task wound analysis under a common protocol. 
WILLIE harmonizes three public wound datasets into a shared benchmark and compares unified models across three scaling configurations against 10 single-task baselines. 
The best model achieves 91.88\% classification accuracy, 91.41\% Dice, and 96.23\% AP@0.5 while producing all three outputs in a single forward pass. 
Beyond aggregate performance, our results show that segmentation-derived localization outperforms dedicated detection baselines in this benchmark, suggesting that box-based localization may be unnecessary for spatially coherent wound targets. 
Our findings highlight that effective multi-task learning in healthcare imaging depends not only on shared representations, but also on task formulation, compatibility, and benchmark design.
\end{abstract}

\section{Introduction}

Chronic wounds remain a major challenge in healthcare delivery \citep{frykberg2015challenges, gould2015chronic, martinengo2019prevalence, olsson2019humanistic}. 
In the United States alone, chronic wound management affects over 8.2 million patients annually and incurs more than \$28 billion in treatment costs \citep{nussbaum2018economic, sen2021}. The Agency for Healthcare Research and Quality (AHRQ) estimates that more than 2.5 million people develop pressure ulcers annually in U.S. hospitals \citep{berlowitz2011preventing}, and updated Medicare claims data indicate that chronic wound prevalence among beneficiaries rose from 14.5\% in 2014 to 16.3\% in 2019 \citep{carter2023chronic}. 
These wounds often require prolonged care, repeated assessment, and frequent treatment adjustment over weeks or months. 
The burden is especially high in settings with limited access to wound specialists, where delays in assessment can lead to infection, hospitalization, amputation, or other complications \citep{armstrong2020five, lavery2006risk, swanson2022iwii, padula2019national, boulton2005global}.

Clinical wound assessment is inherently multi-faceted, demonstrated in Figure~\ref{fig:teaser}.
When evaluating a wound image, clinicians identify wound type, delineate wound boundaries, and localize the wound relative to surrounding anatomy for documentation and longitudinal comparison \citep{berlowitz2011preventing, keast2004measure, sibbald2021optimizing}.  
Wound extent and anatomical context may inform wound type, while accurate boundary delineation is necessary for measuring progression over time. 
These tasks correspond to classification, segmentation, and localization, respectively. 
A computational approach that treats these tasks jointly therefore better reflects the way wound assessment is performed in practice.
Despite this intuitive coupling, \emph{most existing machine learning approaches address wound classification, segmentation, and detection in isolation.} 

In this work, we present \textbf{WILLIE}: \textbf{W}ound \textbf{I}maging, \textbf{L}ocalization, and \textbf{I}ntegrated \textbf{E}valuation, \emph{a unified framework and benchmark for wound classification, segmentation, and localization}. 
We position WILLIE as both a unified wound analysis framework and a common experimental setting for systematically evaluating multi-task wound analysis.
WILLIE integrates three public wound datasets into a shared benchmark and supports direct comparison between unified and single-task models under a common evaluation protocol. 
This setting allows us to examine not only predictive performance, but also broader questions about task compatibility, foundation-model scaling, and when localization should be learned directly versus derived from segmentation.
Our work makes four contributions:
\begin{enumerate}
    \item We introduce a unified wound analysis benchmark by harmonizing three public datasets into a common evaluation setting for classification, segmentation, and localization.
    \item We present WILLIE, a unified framework that supports all three wound-analysis tasks within a single pipeline.
    \item We provide a systematic comparison of unified and single-task models across three scaling configurations and 10 baselines under a shared protocol.
    \item We identify generalizable insights about multi-task learning in healthcare imaging, including task-dependent scaling behavior, the strengths of segmentation-derived localization, and the importance of task compatibility in benchmark design.
\end{enumerate}

\subsection*{Generalizable Insights about Machine Learning in the Context of Healthcare}
This study yields several generalizable insights for healthcare image analysis, multi-task learning, and system design:
\begin{itemize}
\itemsep-0.2em 
    \item For spatially coherent medical targets, accurate segmentation may provide a stronger foundation for localization than a separately trained detection head. In our benchmark, segmentation-derived localization outperforms dedicated detection baselines, suggesting that when the target is contiguous and clinically defined by extent and boundary, mask prediction may be more effective than box regression alone.

    \item The benefits of scaling are task-dependent, even within a single clinical application. In our benchmark, classification performance saturates earlier, whereas segmentation and localization continue to improve with larger pretrained backbones. This suggests model selection should not assume uniform returns from scaling across tasks.

    \item Successful multi-task learning depends not only on shared inputs, but also on compatibility among task objectives. Our results suggest that classification and segmentation benefit from shared learning, whereas localization framed as box regression may be less compatible and may be better derived post hoc.

    \item Residual error can reflect genuine clinical ambiguity rather than purely algorithmic weakness. In our benchmark, persistent confusion between pressure and venous wounds suggests limits of image-only assessment, where visual evidence alone may be insufficient to resolve etiology without additional clinical context. In such settings, models should support uncertainty flagging and escalation to human review rather than present all predictions as equally definitive.
\end{itemize}

\subsection*{Data and Code Availability}
We release the unified and cleaned benchmark developed in this work and the preprocessing pipeline used to construct the benchmarks.
We also provide scripts to reproduce the main results reported in this paper. 
WILLIE benchmark: \url{https://huggingface.co/datasets/QianGroup/willie-benchmark}. Trained models: \url{https://huggingface.co/QianGroup/willie-weights}. Code: \url{https://github.com/Qian-Group-HRI/Willie}

\section{Related Work}
Our work builds on three closely related lines of research: wound image analysis, multi-task learning in medical imaging, and vision foundation models for healthcare.

\subsection{Wound Image Analysis}
Recent wound-image studies have focused primarily on task-specific models for classification, segmentation, or detection.
Deep convolutional networks established strong baselines for wound classification \citep{anisuzzaman2022, aldoulah2023}, while the FUSeg challenge \citep{wang2022} and subsequent U-Net-based methods \citep{dhar2024} standardized evaluation for wound segmentation.
More recent work has explored stronger pretrained backbones, ensemble methods \citep{kiprono2025, ullah2025}, and limited use of clinical metadata \citep{patel2024}.
Wound localization, however, remains less studied and is typically treated as a standalone detection task based on bounding boxes, without integration with diagnostic classification or pixel-level delineation.
Overall, the state of the art remains fragmented across separate tasks and datasets, making it difficult to assess the value of unified wound analysis \citep{alhababi2025}.

\subsection{Multi-Task Learning in Medical Imaging}
Multi-task learning has shown promise in medical imaging by leveraging shared structure across related tasks such as classification, segmentation, and detection.
Prior work has reported gains in representation quality, data efficiency, and regularization \citep{mlynarski2019, esteva2017}, especially in settings where labeled data are limited.
At the same time, successful multi-task learning depends on effective balancing of task objectives \citep{kendall2018, chenZ2018, guo2018}, and existing work has shown that related clinical tasks do not always interact favorably during joint optimization \citep{standley2020which, fifty2021efficiently, yu2020gradient}.
Most prior medical imaging studies focus on two-task settings, with fewer examining three-task combinations under a common evaluation protocol.
Our work extends this literature by studying classification, segmentation, and localization jointly in a clinically grounded wound-analysis benchmark.

\subsection{Vision Foundation Models for Medical Imaging}
Large pretrained vision models are increasingly important in medical imaging.
Self-supervised models such as DINOv2 \citep{oquab2024} provide strong transferable representations for image classification, while ConvNeXt \citep{liu2022convnext} remains competitive through strong local inductive biases.
Segmentation foundation models such as SAM \citep{kirillov2023} and SAM2 \citep{ravi2024} have further advanced mask prediction with limited task-specific adaptation, including medical adaptations such as MedSAM \citep{ma2024}.
These developments are especially relevant to wound analysis, where datasets remain modest in size and tasks differ substantially in visual and structural demands.
Our work builds on these advances by comparing foundation-model backbones within a unified framework and benchmark for wound classification, segmentation, and localization.

\section{The WILLIE Benchmark}

\begin{figure}[t]
  \centering
  \includegraphics[width=0.95\columnwidth]{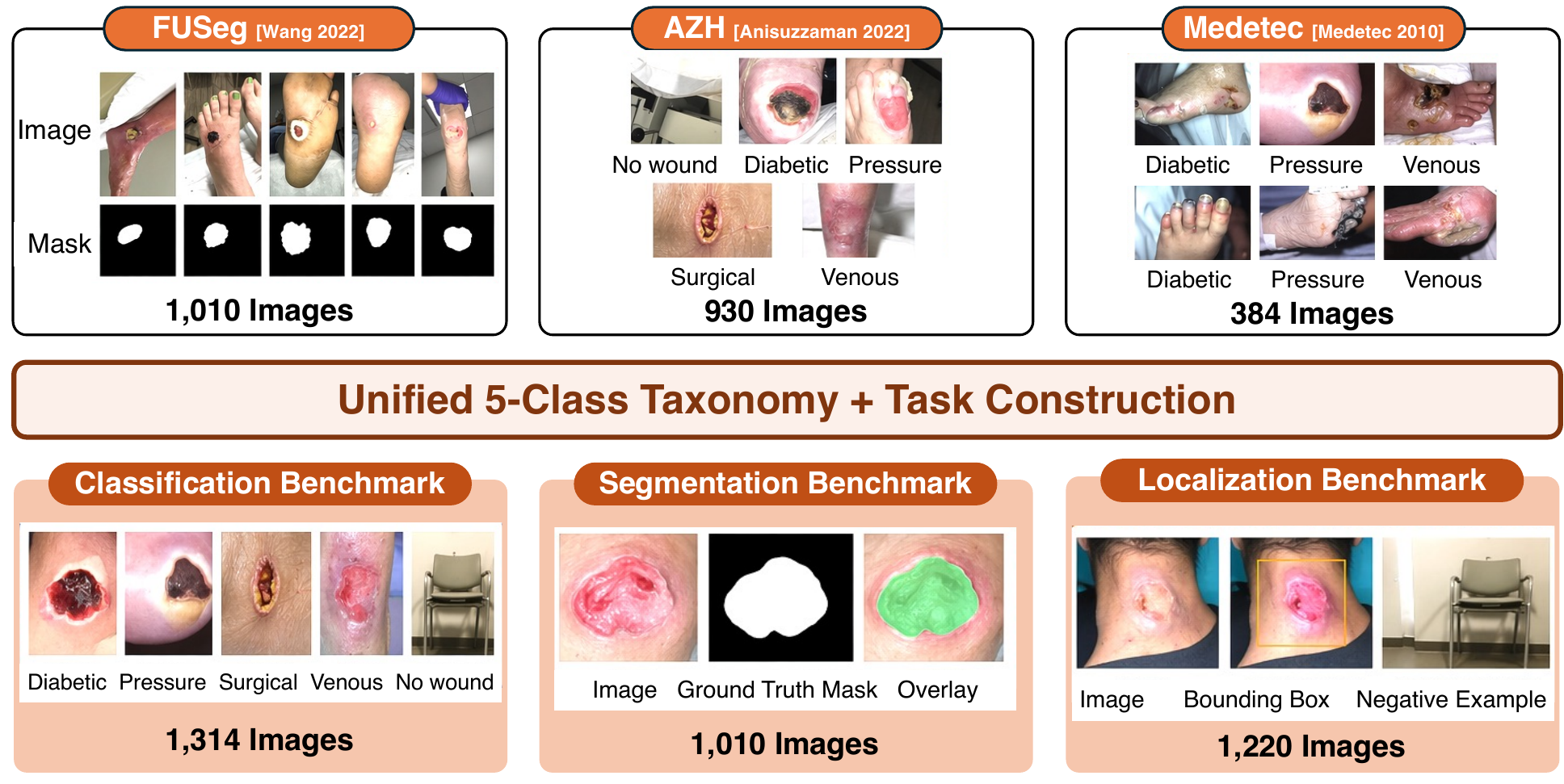}
  \caption{Construction of the WILLIE benchmark from three public wound datasets. FUSeg provides binary masks for segmentation and derived localization, AZH provides expert-annotated wound categories, and Medetec supplements underrepresented wound classes for classification. These sources are harmonized into a unified 5-class taxonomy and shared benchmark supporting classification, segmentation, and localization under a common evaluation protocol.}
  \label{fig:wound-samples}
\end{figure}

A central contribution of this work is the construction of a unified benchmark for wound classification, segmentation, and localization by harmonizing three publicly available datasets: FUSeg, AZH, and Medetec, representative samples shown in Figure~\ref{fig:wound-samples}.
The benchmark approximates real-world wound care diversity: FUSeg provides clinical photographs with segmentation masks, AZH provides expert-annotated classification across wound types, and Medetec supplements underrepresented classes.
Together, they enable evaluation of clinically coupled wound-analysis tasks under a shared experimental framework.

\subsection{Source Datasets}

\paragraph{FUSeg.}
The FUSeg dataset \citep{wang2022} was collected through a collaboration with the Advancing the Zenith of Healthcare (AZH) Wound and Vascular Center and the University of Wisconsin-Milwaukee.
This dataset contains 1,010 chronic wound images with binary segmentation masks at $512 \times 512$ resolution.

\paragraph{AZH.}
The AZH dataset \citep{anisuzzaman2022} 
contains 930 clinical wound photographs annotated for classification across six categories: diabetic (185), pressure (134), surgical (164), venous (247), background (100), and no\_wound (100).
The dataset provides an official split of 696 training images and 234 test images, which we preserve for held-out evaluation.
Image dimensions vary substantially, ranging from $93 \times 93$ to $259 \times 259$ pixels.

\paragraph{Medetec.}
The Medetec dataset \citep{medetec2010} contains 384 free stock images representing diabetic (81), pressure (170), and venous (133).
We use Medetec 
to supplement underrepresented wound classes without introducing cross-dataset evaluation bias.

\subsection{Unified Class Taxonomy}

We unify the classification labels into a 5-class taxonomy: diabetic, pressure, surgical, venous, and no\_wound.
The AZH \texttt{background} category is merged with \texttt{no\_wound}, since both represent the absence of a wound requiring analysis.
This yields a moderately imbalanced classification task, with a maximum class ratio of 2.32:1 (venous to surgical).
To mitigate this imbalance, we use inverse-frequency class weights and weighted sampling during classification training. Table~\ref{tab:class_distribution} summarizes the class distribution across datasets and the inverse-frequency weights used during training.

\begin{table}[t]
\centering
\caption{Classification class distribution with inverse-frequency training weights.}
\resizebox{0.55\columnwidth}{!}{%
\begin{tabular}{lccc|cc}
\toprule
 & \multicolumn{3}{c|}{\textbf{Train}} & \multicolumn{2}{c}{\textbf{Test}} \\
\textbf{Class} & \textbf{AZH} & \textbf{Medetec} & \textbf{Total} & \textbf{AZH} & \textbf{Weight} \\
\midrule
Diabetic  & 139 & 81  & 220 & 46 & 0.982 \\
Pressure  & 100 & 170 & 270 & 34 & 0.802 \\
Surgical  & 122 & --- & 122 & 42 & 1.765 \\
Venous    & 185 & 133 & 318 & 62 & 0.680 \\
No Wound  & 150 & --- & 150 & 50 & 1.434 \\
\midrule
Total     & 696 & 384 & 1,080 & 234 & --- \\
\bottomrule
\end{tabular}}
\label{tab:class_distribution}
\end{table}

\subsection{Task-Specific Benchmark Construction}

\paragraph{Classification.}
The unified classification set contains 1,314 images.
We split these into 918 training images, 162 validation images, and 234 test images using stratified sampling, where the 234-image test set corresponds exactly to the official AZH test split.
For 5-fold cross-validation, the combined training and validation pool of 1,080 images is partitioned into five stratified folds, while the AZH test set remains fixed and held out across all folds.

\paragraph{Segmentation.}
Segmentation experiments use the official FUSeg release, comprising 610 training images and 400 validation images with binary masks at $512 \times 512$ resolution.
The official FUSeg test set is excluded because its ground truth is not publicly available.

\paragraph{Localization.}
Localization annotations are derived from FUSeg segmentation masks by converting connected wound regions into bounding boxes, and are supplemented with negative images from AZH.
This yields 1,220 localization images: 810 for training (600 positive and 210 negative) and 400 for validation (386 positive and 14 negative).

\subsection{Preprocessing and Augmentation}

We consider three model scales throughout the benchmark: \textbf{MINI}, \textbf{BASE}, and \textbf{XL}.
These variants differ in model capacity and are used to examine the effect of scaling across classification, segmentation, and localization tasks.
More details are provided in the next section.

For classification, the \textbf{MINI} configuration operates on images resized to $224 \times 224$, whereas the \textbf{BASE} and \textbf{XL} configurations use $384 \times 384$ inputs.
Training-time augmentation includes random horizontal flips ($p=0.5$), random vertical flips ($p=0.3$), random rotations within $\pm 20^\circ$, color jitter (brightness $=0.2$, contrast $=0.2$, saturation $=0.1$, hue $=0.05$), and random erasing ($p=0.1$).
All images are normalized using ImageNet statistics.

For segmentation, images and masks are processed at $512 \times 512$ resolution.
We apply the same spatial augmentations as in classification, with nearest-neighbor interpolation for masks to preserve discrete labels.

For localization, images are resized to $512 \times 512$ and augmented using 
mosaic augmentation and HSV perturbation.
When bounding boxes are derived from segmentation masks, we use nearest-neighbor interpolation during resizing to maintain spatial consistency.

\section{The WILLIE Framework}

\begin{figure}[t]
  \centering
  \includegraphics[width=0.95\columnwidth]{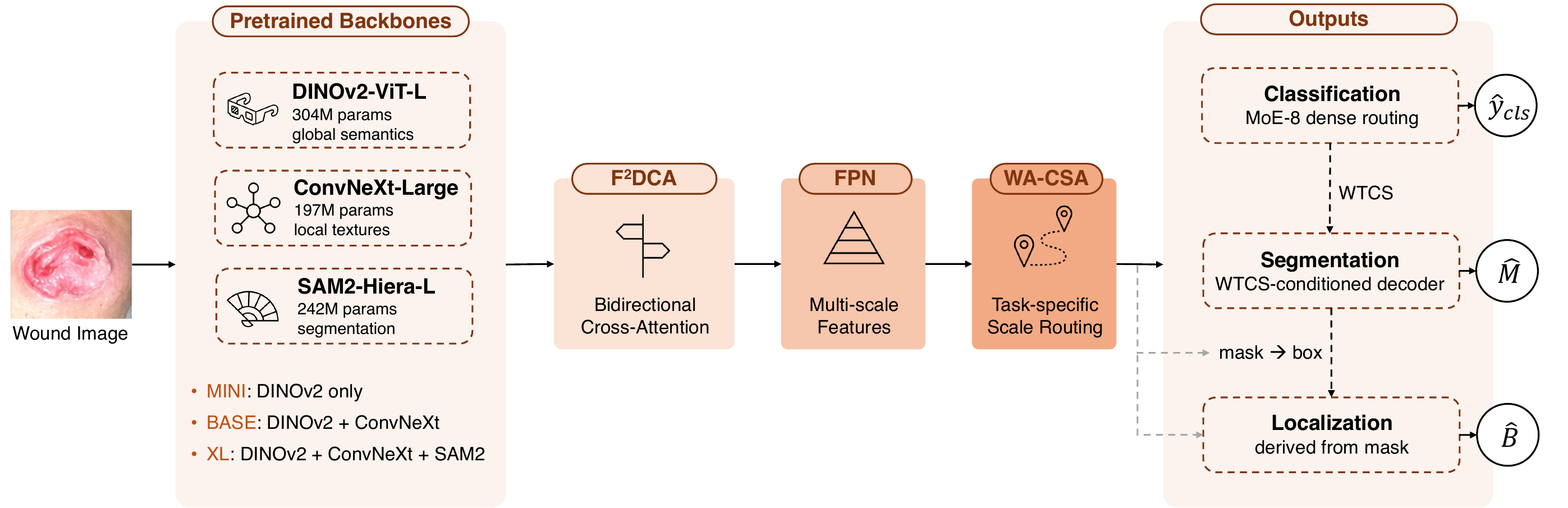}
  \vspace{-1em}
  \caption{Overview of WILLIE architecture. A wound image is processed through shared pretrained backbones, cross-backbone fusion, multi-scale feature construction, and task-specific scale routing to produce three outputs: wound type classification, boundary segmentation, and localization. }
  \label{fig:willie-architecture}
\vspace{-1em}
\end{figure}

WILLIE is a unified framework for wound classification, segmentation, and localization.
As demonstrated in Figure~\ref{fig:willie-architecture}, given an input wound image $\mathbf{x}$, WILLIE produces three outputs: a wound-type prediction $\hat{y}_{\text{cls}}$, a segmentation mask $\hat{\mathbf{M}}$, and a localization result $\hat{\mathbf{B}}$ derived from the predicted mask.
The framework is evaluated in three model scales: \textbf{MINI}, \textbf{BASE}, and \textbf{XL}, to study how backbone capacity affects multi-task performance under a common benchmark.
Our emphasis is not only on model design, but also on using a shared framework to evaluate when joint modeling is effective across clinically coupled wound-analysis tasks.

\subsection{Model Scales and Backbone Configurations}

We consider three model scales with increasing backbone capacity, summarized by Table~\ref{tab:param_stats}.

\begin{table}[H]
\vspace{-1em}
\centering
\caption{Model variant parameter statistics.}
\resizebox{0.9\columnwidth}{!}{%
\begin{tabular}{llcccc}
\toprule
\textbf{Variant} & \textbf{Backbone(s)} & \textbf{Total} & \textbf{Trainable} & \textbf{Frozen} & \textbf{\% Trainable} \\
\midrule
\textbf{MINI} & DINOv2-S & 34.3M & 12.3M & 22.1M & 35.7 \\
\textbf{BASE} & DINOv2-L + ConvNeXt-L & 520.4M & 19.8M & 500.6M & 3.8 \\
\textbf{XL}   & DINOv2-L + ConvNeXt-L + SAM2-Hiera-L & 762.5M & 360.2M & 402.3M & 47.3 \\
\bottomrule
\end{tabular}}
\label{tab:param_stats}
\end{table}

\paragraph{WILLIE-MINI, 34.3M parameters.}
MINI uses a single DINOv2-Small \citep{oquab2024} (ViT-S/14) backbone with 22.06M frozen parameters and 12.26M trainable parameters in the task heads. 
It omits $F^2$DCA and uses a simplified feature pyramid to upsample the single-backbone output to multiple scales. 
This lightweight variant serves as the lowest-capacity point in our scaling analysis and a resource-efficient configuration.

\paragraph{WILLIE-BASE, 520.4M parameters.}
BASE uses two pretrained backbones: DINOv2-ViT-Large (304M) for global semantics and ConvNeXt-Large \citep{liu2022convnext} (197M) for local texture and boundary cues. Only 19.8M parameters (3.8\%) are trainable, including the $F^2$DCA layers, feature pyramid, task heads, and WTCS parameters. This design enables complementary feature extraction while limiting overfitting on a modest-sized benchmark.

\paragraph{WILLIE-XL, 762.5M parameters.}
XL adds SAM2-Hiera-Large \citep{ravi2024} (242M) as a third backbone, forming a triple-backbone architecture. 
DINOv2 and ConvNeXt remain frozen, while selected SAM2 decoder layers are fine-tuned, yielding 360.2M trainable parameters (47.3\%). 
This variant tests whether a larger, segmentation-oriented backbone primarily benefits dense prediction rather than all wound-analysis tasks uniformly.

\subsection{Shared Representation and Feature Fusion}
WILLIE combines multi-backbone configurations through cross-backbone and cross-scale fusion.
Let $F_A$ and $F_B$ denote feature maps from two backbones at a given scale.
$F^2$DCA addresses the semantic gap between these representations through bidirectional cross-attention:
\[
F_A' = F_A + \alpha \cdot \mathrm{WoundGate}(F_A)\odot \mathrm{CrossAttn}(F_A,F_B)
\]
with an analogous update for $F_B'$.
Here, $\mathrm{WoundGate}(F)=\sigma(W_{\mathrm{gate}}\cdot \mathrm{GAP}(F))$ is a learned sigmoid gating function, where GAP denotes global average pooling.
The gate suppresses cross-attention in non-wound regions and amplifies it where complementary features are most useful.
The scaling factor $\alpha$ is initialized to 0.1 and typically converges to 0.15--0.25 during training, so that cross-attention corrections remain a bounded refinement rather than dominating the original backbone features.

For XL, $F^2$DCA is applied pairwise across all three backbone combinations at each scale.
The resulting enhanced features are merged by learned weighted summation within the feature pyramid \citep{lin2017}.
At scale $s$, the fused pyramid feature is
\[
\mathrm{FPN}_s = \sum_b \alpha_b^s \cdot \mathrm{Proj}_b^s(F_b^s),
\qquad
\sum_b \alpha_b^s = 1,
\vspace{-0.5em}
\]
where $\mathrm{Proj}_b^s(\cdot)$ denotes a $1\times1$ projection to the common pyramid dimension $d_{\mathrm{fpn}}=256$.
The weights $\alpha_b^s$ are produced by a gating network,
\[
\alpha^s = \mathrm{softmax}\!\Big(W^s \cdot [\mathrm{GAP}(F_1^s);\mathrm{GAP}(F_2^s);\ldots]\Big).
\vspace{-0.5em}
\]
This allows the framework to emphasize different backbones at different scales.
Empirically, higher-resolution levels place greater weight on ConvNeXt for local boundary detail, lower-resolution levels place greater weight on DINOv2 for global wound-type structure, and intermediate levels place greater weight on SAM2 in the XL model.

To support task-specific use of the shared pyramid, WILLIE applies Wound-Aware Cross-Scale Attention (WA-CSA).
Given a task query $\mathbf{q}$ and pyramid features $\mathbf{P}=\{P_1,P_2,P_3,P_4\}$, WA-CSA computes
\[
\mathrm{WA\text{-}CSA}(\mathbf{q},\mathbf{P})=\sum_j w_j(\mathbf{q})\cdot \mathrm{Attn}(\mathbf{q},P_j),
\vspace{-0.5em}
\]
where $w_j(\mathbf{q})=\mathrm{softmax}_j(\mathrm{MLP}(\mathbf{q}))$ is a learned per-query, per-scale weighting.
This mechanism allows wound boundary queries to emphasize higher-resolution levels and wound-type queries to emphasize lower-resolution levels.

\subsection{Task Heads}
\label{subsec:task-heads}

\paragraph{Classification.}
The classification branch applies global pooling to the shared representation and predicts one of five wound classes using a mixture-of-experts classifier with eight expert subnetworks (MoE-8) \citep{jacobs1991}.\revised{~Per-variant expert counts are detailed in Appendix~\ref{app:moe}.}
Let $\mathbf{h}_{\text{cls}}$ be the pooled classification feature; each expert is a two-layer MLP with hidden dimension 512:
\[
\mathrm{Expert}_k(\mathbf{h}_{\text{cls}}) = W_2^k\,\mathrm{GELU}(W_1^k\mathbf{h}_{\text{cls}}+b_1^k)+b_2^k.
\]
A router produces soft weights
$\mathbf{w}=\mathrm{softmax}(W_{\mathrm{router}}\mathbf{h}_{\text{cls}})$,
and the final classification logits are
$\mathbf{o}_{\text{cls}} = \sum_{k=1}^{8} w_k \cdot \mathrm{Expert}_k(\mathbf{h}_{\text{cls}})$.
Then the classification result: 
\[
\hat{y}_{\text{cls}} = \arg\max \mathbf{o}_{\text{cls}}.
\]
WILLIE uses dense routing, so all experts contribute to every prediction.
An auxiliary load-balancing loss encourages more even expert utilization and reduces expert collapse.

\paragraph{Segmentation.}
The segmentation branch follows a U-Net decoder \citep{ronneberger2015} with projected spatial and channel squeeze-excitation (P-scSE) blocks for feature recalibration at each decoder level.
To couple diagnostic context with boundary prediction, the decoder is modulated by Wound-Type Conditioned Segmentation (WTCS), implemented with feature-wise linear modulation (FiLM) \citep{perez2018}.
At decoder level $\ell$, the penultimate classification feature vector $\mathbf{z}$ computes a per-channel scale and shift:
\[
\mathrm{WTCS}(F_\ell,\mathbf{z})=\gamma_\ell \odot F_\ell + \beta_\ell,
\]
where $\gamma_\ell = W_\gamma^\ell \mathbf{z} + b_\gamma^\ell$ and $\beta_\ell = W_\beta^\ell \mathbf{z} + b_\beta^\ell$.
This conditioning allows wound-type information to influence mask prediction without requiring hard class assignments, while preserving gradient flow between the segmentation and classification branches.
The modulated decoder features are then mapped to the predicted wound mask $\hat{\mathbf{M}}$:
\[
\hat{\mathbf{M}} = \mathrm{Decoder}_{\text{seg}}(\{\,\mathrm{WTCS}(F_\ell,\mathbf{z})\,\}_\ell).
\]

\paragraph{Localization.}
Rather than training a separate detection head, WILLIE derives wound localization from the predicted segmentation mask.
The segmentation-to-localization pipeline thresholds the predicted mask $\hat{\mathbf{M}}$ at 0.5, extracts connected components, computes one axis-aligned bounding box per component, and removes components smaller than 20 pixels, yielding the localization output $\hat{\mathbf{B}}$.\revised{~Full construction details are provided in Appendix~\ref{app:loc}.}
This produces localization as a derived output of segmentation rather than as an independently trained regression task.

\emph{Why localization is derived from segmentation.}
We treat localization as a derived output rather than a separately trained detection objective for both clinical and empirical reasons.
In wound images, the target region is typically contiguous and defined by its extent and boundary, making segmentation a natural basis for localization.
In preliminary experiments, a dedicated FCOS head \citep{tian2019} achieved only 6.6\% mAP@0.5 within the shared architecture, whereas segmentation-derived localization performed substantially better.
This design also avoids introducing a separate detection loss into the multi-task objective.
Within the benchmark, localization is therefore evaluated as an output of the unified framework, while standard detection baselines are retained as external references.

\subsection{Training Objective}
\label{subsec:training-objective}

WILLIE is trained with a composite multi-task loss:
\[
\mathcal{L}_{\mathrm{total}}
=
\lambda_{\mathrm{cls}}\mathcal{L}_{\mathrm{cls}}
+
\lambda_{\mathrm{seg}}\mathcal{L}_{\mathrm{seg}}
+
\lambda_{\mathrm{bal}}\mathcal{L}_{\mathrm{balance}},
\]
where $\mathcal{L}_{\mathrm{cls}}$ is weighted cross-entropy computed from $\hat{y}_{\mathrm{cls}}$, $\mathcal{L}_{\mathrm{seg}}$ is the equally weighted sum of binary cross-entropy and Dice loss computed from $\hat{\mathbf{M}}$,
$
\mathcal{L}_{\mathrm{seg}} = 0.5\,\mathcal{L}_{\mathrm{BCE}} + 0.5\,\mathcal{L}_{\mathrm{Dice}},
$
and $\mathcal{L}_{\mathrm{balance}}$ is the MoE load-balancing regularizer.
The task weights $\lambda_{\mathrm{cls}}$ and $\lambda_{\mathrm{seg}}$ are learnable parameters initialized to 1.0.

Since localization is derived from segmentation rather than trained as a separate head, there is no localization loss.
For checkpoint selection, we use a combined validation score:
\[
\mathrm{Combined}
=
0.4 \times \mathrm{cls\_acc}
+
0.4 \times \mathrm{seg\_dice}
+
0.2 \times \mathrm{loc\_AP@0.5},
\]
where $\mathrm{loc\_AP@0.5}$ is computed from the seg-to-loc pipeline on the validation set.
This criterion reflects the central goal of the benchmark: evaluating unified performance across clinically coupled wound-analysis tasks rather than optimizing any one task in isolation.

\section{Experiments}
We evaluate WILLIE under a common benchmark protocol and compare it against established task-specific baselines for wound classification, segmentation, and localization.

\subsection{Hypotheses}

We posit the following hypotheses.

\paragraph{H1:} A unified multi-task framework can match or exceed strong task-specific baselines while producing all three outputs in a single pipeline. 
\textbf{H1} will be evaluated by comparing WILLIE against task-specific baselines on classification accuracy, macro F1, and macro AUC-ROC; segmentation Dice and IoU; and localization AP@0.5.

\paragraph{H2:} Localization derived from segmentation is more effective than training a separate detection head or relying solely on standard detection architectures. 
\textbf{H2} will be evaluated by comparing WILLIE's segmentation-derived localization performance against baselines, and against preliminary experiments with a dedicated FCOS detection head.

\paragraph{H3:} The benefits of scaling are task-dependent: larger and more specialized pretrained backbones provide larger gains for segmentation and localization than for classification. 
\textbf{H3} will be evaluated by comparing task-specific performance across the MINI, BASE, and XL variants of WILLIE using classification, segmentation, and localization metrics.

\paragraph{H4:} Joint modeling is most effective for task combinations with compatible objectives; in particular, classification and segmentation benefit from shared representations, whereas localization is better treated as a derived output from segmentation than as a third jointly optimized regression task. 
\textbf{H4} will be evaluated by comparing unified and task-specific performance, and by examining whether segmentation-derived localization is more effective than adding a separately optimized detection objective within the shared framework.

\subsection{Baselines}
We compare WILLIE against established baselines in each task, all trained under our unified 5-class protocol with identical data splits and evaluation to ensure fair comparison. 

\paragraph{Classification:}
\begin{itemize}
\vspace{-0.5em}
\itemsep-0.1em 
    \item \textbf{VGG-19} (124.9M) is included following \citet{anisuzzaman2022}.
    
    \item \textbf{ResNet-50} (24.6M) serves as a standard CNN baseline.
    
    \item \textbf{EfficientNet-B4} (18.5M) provides a stronger CNN baseline with recent EfficientNet-based classifiers \citep{aldoulah2023,ullah2025}.
    
    \item \textbf{DINOv2 + Linear} (22.2M) uses a frozen DINOv2 with a linear head, isolating the contribution of self-supervised pretrained features without multi-task modeling.
\end{itemize}

\paragraph{Segmentation:}
\begin{itemize}
\vspace{-0.5em}
\itemsep-0.1em 
    \item \textbf{U-Net with EfficientNet-B3 encoder} (13.2M parameters) provides a standard medical-image segmentation baseline \citep{ronneberger2015}.

    \item \textbf{MedSAM (ViT-B)} (93.7M total, 4.1M trainable) represents a medical adaptation of SAM using prompt-free inference \citep{ma2024}.

    \item \textbf{SAM2.1 (Hiera-S)} (46.1M total, 11.7M trainable) represents a more recent segmentation foundation model \citep{ravi2024}.
\end{itemize}
 We compare SAM2.1 and MedSAM using a Wilcoxon signed-rank test, with $p < 0.001$, indicating a statistically significant advantage for SAM2.1 on this benchmark.

\paragraph{Localization:}
\begin{itemize}
\vspace{-0.5em}
\itemsep-0.1em 
    \item \textbf{RT-DETR-L} (32.8M parameters) provides a transformer-based localization baseline on bounding boxes derived from FUSeg masks \citep{zhao2024} .

    \item \textbf{YOLOv8m} (25.9M parameters) provides a strong real-time detection baseline trained under the same conditions \citep{jocher2023}.
\end{itemize}

\subsection{Training Protocol}
All WILLIE variants are trained using 5-fold stratified cross-validation on the 1,080-image training and validation pool, with approximately 864 images for training and 216 for validation in each fold.
The official 234-image AZH test split is held out across all folds and used only for final classification evaluation.

For the optimizer, we use AdamW with a weight decay of 0.01.
Learning rates are set to $1.5\times10^{-5}$ for fine-tuned backbone components and $1\times10^{-4}$ for task-specific heads.
Training uses cosine annealing with warm restarts ($T_0=10$, $T_{\mathrm{mult}}=1$), for a maximum of 50 epochs, with early stopping based on the combined validation metric (patience $=12$).

Batch size is 8 for MINI and 16 for BASE and XL.
Mixed-precision training is used throughout.
We additionally apply exponential moving average model averaging with decay 0.999.
All experiments are run on NVIDIA Tesla V100-PCIE-32GB GPUs.\revised{~Latency and peak-memory characterization is reported in Appendix~\ref{app:latency}.}

\subsection{Evaluation Metrics}

We report task-specific and unified metrics chosen to test the hypotheses above.

\paragraph{Classification.}
We report accuracy, macro F1, and macro AUC-ROC using a one-vs-rest formulation.
These metrics are used to evaluate \textbf{H1}.
Macro F1 and macro AUC-ROC are particularly important because the unified 5-class setting is moderately imbalanced.

\paragraph{Segmentation.}
We report Dice coefficient and intersection-over-union (IoU), averaged across images.
These metrics are used to evaluate both \textbf{H1} and \textbf{H3}.
They test whether unified modeling preserves strong boundary delineation relative to task-specific segmentation baselines, and whether segmentation continues to benefit from larger and more specialized pretrained backbones.

\paragraph{Localization.}
We report AP@0.5.
For baselines, AP@0.5 is computed from detector outputs.
For WILLIE, AP@0.5 is computed from the segmentation-to-localization pipeline.
This metric is used to evaluate \textbf{H1} and \textbf{H2} and to test whether segmentation-derived localization is an effective alternative to dedicated detection.

\paragraph{Unified evaluation.}
To assess overall multi-task performance, we use the combined validation score defined in Section~\ref{subsec:training-objective}.

\subsection{Results}

We report results using the evaluation protocol described above.

\begin{table}[t]
\centering
\caption{Classification comparison on the held-out AZH test set ($n=234$).}
\label{tab:cls_results}
\resizebox{0.75\columnwidth}{!}{%
\begin{tabular}{llcccc}
\toprule
\textbf{Model} & \textbf{Type} & \textbf{Params} & \textbf{Acc (\%)} & \textbf{F1 (\%)} & \textbf{AUC} \\
\midrule
\textbf{VGG-19}                  & Baseline & 124.9M & 78.63 & 78.14 & 0.956 \\
\textbf{EfficientNet-B4}         & Baseline & 18.5M  & 83.76 & 83.36 & 0.975 \\
\textbf{DINOv2+Linear}           & Ablation & 22.2M  & 86.75 & 87.01 & 0.986 \\
\textbf{WILLIE-MINI (ens)}       & Unified  & 34.3M  & 88.90 & 87.60 & 0.976 \\
\textbf{ResNet-50}               & Baseline & 24.6M  & 88.03 & 87.82 & 0.973 \\
\textbf{WILLIE-XL (TTA)}         & Unified  & 762.5M & \textbf{91.88} & 90.73 & 0.986 \\
\textbf{WILLIE-BASE (TTA)}       & Unified  & 520.4M & \textbf{91.88} & \textbf{91.14} & \textbf{0.992} \\
\bottomrule
\end{tabular}}
\end{table}

\paragraph{Result 1: Unified approaches outperform or perform comparably relative to task-specific baselines.}
Table~\ref{tab:cls_results} shows that the unified WILLIE models achieve the strongest classification performance. BASE and XL attain the highest accuracy, and BASE achieves the strongest overall macro-F1 and AUC, indicating that unified modeling does not compromise diagnostic classification performance. Notably, even the lightweight MINI remains competitive with the strongest baseline. These results support H1 for classification.

\begin{table}[t]
\centering
\caption{Segmentation comparison on FUSeg validation ($n=400$).}
\label{tab:seg_results}
\resizebox{0.7\columnwidth}{!}{%
\begin{tabular}{llccc}
\toprule
\textbf{Model} & \textbf{Type} & \textbf{Params} & \textbf{Mean Dice} & \textbf{Median Dice} \\
\midrule
\textbf{U-Net (Eff-B3)}   & Baseline & 13.2M  & 84.59 & 91.20 \\
\textbf{WILLIE-MINI}      & Unified  & 34.3M  & 81.80 & 87.50 \\
\textbf{WILLIE-BASE}      & Unified  & 520.4M & 86.36 & 92.89 \\
\textbf{MedSAM (ViT-B)}   & Baseline & 93.7M  & 90.93 & 93.33 \\
\textbf{WILLIE-XL}        & Unified  & 762.5M & 91.41 & 93.50 \\
\textbf{SAM2.1 (Hiera-S)} & Baseline & 46.1M  & \textbf{91.74} & \textbf{93.76} \\
\bottomrule
\end{tabular}}
\end{table}

Table~\ref{tab:seg_results} shows that segmentation performance improves substantially with model scale and backbone specialization. XL achieves segmentation performance comparable to the strongest task-specific baselines, exceeding both U-Net and MedSAM and approaching SAM2.1. In contrast, the lower-capacity unified variants lag behind the strongest segmentation specialists, suggesting that dense wound delineation places greater demands on representation quality than classification. These results support H1 for segmentation. 

\begin{table}[t]
\centering
\caption{Localization comparison on FUSeg validation ($n=400$).}
\label{tab:loc_results}
\resizebox{0.75\columnwidth}{!}{%
\begin{tabular}{llccccc}
\toprule
\textbf{Model} & \textbf{Type} & \textbf{Params} & \textbf{AP@0.5} & \textbf{Prec.} & \textbf{Recall} & \textbf{F1} \\
\midrule
\textbf{RT-DETR-L}              & Dedicated & 32.8M  & 87.95 & 88.15 & 89.38 & 88.76 \\
\textbf{YOLOv8m}                & Dedicated & 25.9M  & 91.22 & 87.45 & 88.47 & 87.96 \\
\textbf{WILLIE-MINI s$\to$l}    & Unified   & 34.3M  & 84.60 & 89.03 & 88.66 & 88.84 \\
\textbf{WILLIE-BASE s$\to$l}    & Unified   & 520.4M & 89.91 & 92.40 & 95.40 & 93.87 \\
\textbf{WILLIE-XL s$\to$l}      & Unified   & 762.5M & \textbf{96.23} & \textbf{96.23} & \textbf{94.64} & \textbf{95.43} \\
\bottomrule
\multicolumn{7}{l}{\scriptsize s$\rightarrow$l: segmentation-to-localization pipeline.}
\end{tabular}}
\vspace{-1em}
\end{table}

Table~\ref{tab:loc_results} shows that segmentation-derived localization is highly effective. XL achieves the strongest localization performance, surpassing YOLOv8m and RT-DETR-L, and BASE remains competitive with dedicated detectors despite not optimizing a separate box-regression objective. 
These results support H1 for localization and directly support H2.

\begin{figure}[b]
\vspace{-1em}
    \centering
    \includegraphics[width=\textwidth]{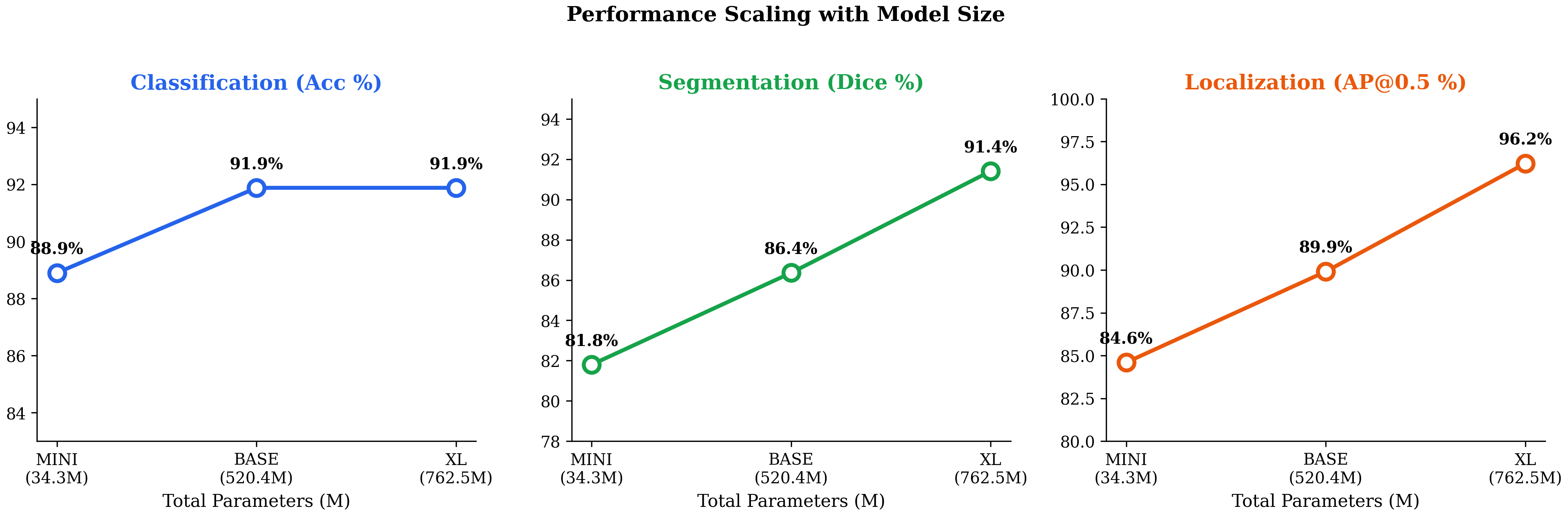}
    \vspace{-1.5em}
    \caption{Task-dependent scaling behavior across WILLIE variants. Classification accuracy improves from MINI to BASE and then saturates at XL, whereas segmentation Dice and localization AP@0.5 continue to improve with model scale. 
    }
    \label{fig:scaling}
    \vspace{-1.5em}
\end{figure}

\paragraph{Result 2: Scaling benefits are strongly task-dependent, with the largest gains observed for segmentation and localization.}
Figure~\ref{fig:scaling} shows that scaling behavior is strongly task-dependent. Classification improves from MINI to BASE but does not improve further at XL, indicating early saturation for wound-type recognition under the current benchmark. By contrast, segmentation and localization continue to improve across all three configurations, with the largest gains observed for the dense prediction tasks. This pattern suggests that increased model capacity and backbone specialization provide their clearest benefit for segmentation and segmentation-derived localization, whereas classification appears to benefit primarily from the transition from the lightweight MINI configuration to the stronger BASE configuration. These results support H3.

\paragraph{Result 3: Unified modeling is most effective when task objectives are compatible, with localization best handled as a derived output of segmentation.}
The strongest localization results are achieved when localization is treated as a derived output of segmentation rather than as a separately optimized detection objective, suggesting that these tasks are not equally compatible under joint optimization. In particular, classification and segmentation appear to benefit from shared learning, whereas localization framed as box regression is better handled post hoc in this benchmark. This result supports H4.

\begin{table}
\centering
\caption{Per-class results for WILLIE-BASE (top-3 TTA).}
\label{tab:per_class_results}
\resizebox{0.55\columnwidth}{!}{%
\begin{tabular}{lccccc}
\toprule
\textbf{Class} & \textbf{Prec.} & \textbf{Recall} & \textbf{F1} & \textbf{Support} & \textbf{Train} \\
\midrule
\textbf{Diabetic}  & 97.3 & 78.3  & 86.7 & 46 & 220 \\
\textbf{Pressure}  & 76.5 & 76.5  & 76.5 & 34 & 270 \\
\textbf{Surgical}  & 88.9 & 95.2  & 92.0 & 42 & 122 \\
\textbf{Venous}    & 91.0 & 98.4  & 94.6 & 62 & 318 \\
\textbf{No Wound}  & 98.0 & 100.0 & \textbf{99.0} & 50 & 150 \\
\bottomrule
\end{tabular}}
\end{table}

\paragraph{Result 4: Residual classification error is concentrated in clinically challenging wound categories rather than uniformly distributed across classes.}
Table~\ref{tab:per_class_results} shows that classification performance is strongest for no wound, venous, and surgical categories, while pressure wounds remain the most challenging class despite substantial training representation. The dominant residual error pattern is confusion between pressure and venous wounds, localizing the primary source of remaining classification error within the benchmark. This pattern suggests that some residual error may reflect genuine clinical ambiguity in image-only wound assessment rather than uniformly distributed model weakness.\revised{~Bootstrap 95\% confidence intervals for per-class F1 are reported in Appendix~\ref{app:ci}, and a source-level analysis of the pressure--venous confusion is provided in Appendix~\ref{app:pv}.}

\section{Discussion}
\label{sec:discussion-ref}

Across tasks, WILLIE remained competitive with strong task-specific baselines while revealing broader patterns about task formulation, scaling, and residual error. These findings are relevant not only to wound analysis, but also to healthcare imaging settings in which multiple clinically coupled outputs must be generated from the same input.

\subsection{Key Findings}
\paragraph{Finding 1: Segmentation-derived localization is well suited to wound analysis.}
One of our clearest findings is that localization derived from segmentation can outperform dedicated detection baselines in wound images. 
WILLIE-XL achieved the strongest localization performance overall, exceeding YOLOv8m and RT-DETR-L, whereas a preliminary FCOS head performed poorly within the shared framework. 
This is consistent with the clinical structure of wounds: unlike many generic detection targets, wounds are contiguous regions whose extent and boundary are central to interpretation. 
More broadly, this result suggests that for healthcare imaging tasks in which the target is spatially coherent and clinically defined by region extent, localization may be more effective as a derived output of segmentation than as an independently optimized objective. 

\paragraph{Finding 2: Scaling benefits are strongly task-dependent.}
A second finding is that scaling does not benefit all wound-analysis tasks equally. 
Classification improved from MINI to BASE but then saturated at XL, whereas segmentation and localization continued to improve at the largest scale. 
This indicates that model capacity and backbone specialization matter differently across tasks, even within a single clinical workflow.
For classification, the combination of DINOv2 and ConvNeXt in BASE appears sufficient to capture most of the image's discriminative information. 
For segmentation and localization, the additional features introduced by SAM2 continue to provide measurable gains. 
More generally, these results suggest that scaling decisions in healthcare imaging should be guided by task requirements rather than by the assumption that larger models improve all outputs uniformly.

\paragraph{Finding 3: Unified performance depends on task compatibility.}
The results further suggest that strong unified performance depends not only on feature sharing, but also on whether task objectives remain compatible within a shared architecture. 
In WILLIE, classification and segmentation can be learned jointly while preserving strong performance on both tasks. Localization is also effective within the unified framework, but only when treated as a derived output of segmentation; when framed as a separately optimized detection objective, performance deteriorates sharply.
This pattern highlights an important principle for healthcare multi-task learning: clinically related tasks are not necessarily equally compatible during optimization. 
A unified model may succeed when tasks share useful structure, but degrade when one objective is poorly aligned with the others. 

\paragraph{Finding 4: Residual error reflects clinically meaningful ambiguity.}
Although WILLIE achieves strong overall classification performance, pressure wounds remain the most difficult class, and confusion between pressure and venous wounds is the dominant residual error pattern. This is clinically plausible, as visual appearance alone may not always provide enough information to distinguish etiology when morphology overlaps or when images lack contextual information such as anatomical site, patient history, vascular assessment, or temporal progression.
This observation highlights an important limit of image-only prediction. In healthcare imaging, some residual error may reflect ambiguity in the modality itself rather than insufficient model capacity alone. For wound analysis, this suggests that future systems may benefit from integrating images with metadata, longitudinal evidence, or clinician-in-the-loop escalation rather than treating image-only classification as fully self-sufficient.

\subsection{Limitations}
\label{sec:limitations-ref}

This study has several limitations. First, the benchmark is constructed from a relatively small number of public wound datasets, and the resulting taxonomy remains coarse relative to the heterogeneity of wound care in practice. Second, segmentation evaluation is limited to the public FUSeg training/validation release because official test-set labels are not available. Third, localization is derived from segmentation masks and assumes connected wound regions, which may be less appropriate for images containing multiple disjoint wounds or more complex spatial structure. Fourth, although the framework contains several architectural components, \revised{we provide a per-component ablation in Appendix~\ref{app:ablation} rather than a full factorial ablation of every module}. Finally, we do not provide prospective clinical validation, workflow evaluation, or latency characterization for real-world deployment.

\subsection{Future Work}

Several directions follow naturally from this work. One priority is to extend the benchmark with larger and more diverse wound datasets, including broader wound categories, richer metadata, and images collected across institutions, devices, and care settings. A second is to evaluate unified wound analysis prospectively in realistic clinical workflows, including how model outputs support longitudinal monitoring, triage, and specialist escalation. A third is to study more explicitly which task combinations are compatible in shared healthcare imaging systems, including whether compatibility can be predicted or improved through alternative objective design. Finally, the lightweight MINI configuration suggests a possible path toward lower-resource deployment, but this requires dedicated study of efficiency, robustness, and clinical usability beyond retrospective benchmark evaluation.

\section{Conclusion}

We presented WILLIE, a unified framework and benchmark for wound classification, segmentation, and localization that enables systematic study of multi-task wound analysis under a shared, clinically grounded protocol. 
We show that a single framework can achieve strong performance across all three tasks while producing unified outputs in one pipeline. 
Beyond aggregate performance, the benchmark yields broader insights for healthcare imaging: segmentation-derived localization outperforms dedicated detection baselines in this setting, scaling benefits are strongly task-dependent, and successful unified learning depends not only on shared representations but also on compatibility among task formulations. 

\newpage
\bibliography{main}

\newpage
\appendix

\newpage
\section{Component Ablation Study}
\label{app:ablation}

We ablate each component (F\textsuperscript{2}DCA, WA-CSA, MoE, WTCS) individually by retraining under the same 5-fold cross-validation protocol used in the main paper and evaluating on the validation folds. We report classification accuracy, segmentation Dice, and localization AP@0.5 (mask-derived). No test-time augmentation is used, so absolute numbers are lower than the TTA test-set results in Tables~\ref{tab:cls_results}--\ref{tab:loc_results}; this ablation isolates the \emph{relative} contribution of each component. MINI omits F\textsuperscript{2}DCA by design, as it uses a single backbone.

\begin{table}[h]
\centering
\caption{Components ablation: classification accuracy (\%). 5-fold CV, validation folds, no TTA.}
\label{tab:abl-cls}
\begin{tabular}{lccc}
\toprule
Configuration & MINI & BASE & XL \\
\midrule
Full              & 86.6 & 88.2 & 89.8 \\
$-$ F\textsuperscript{2}DCA & --   & 87.8 & 89.9 \\
$-$ WA-CSA        & 86.6 & 86.9 & 90.2 \\
$-$ MoE           & 86.7 & 87.6 & 89.8 \\
$-$ WTCS          & 86.1 & 87.0 & 89.5 \\
\bottomrule
\end{tabular}
\end{table}

\begin{table}[h]
\centering
\caption{Component ablation: segmentation Dice (\%). 5-fold CV, validation folds, no TTA.}
\label{tab:abl-seg}
\begin{tabular}{lccc}
\toprule
Configuration & MINI & BASE & XL \\
\midrule
Full              & 82.3 & 84.1 & 83.2 \\
$-$ F\textsuperscript{2}DCA & --   & 84.9 & 83.6 \\
$-$ WA-CSA        & 82.9 & 84.7 & 84.0 \\
$-$ MoE           & 83.5 & 83.8 & 82.1 \\
$-$ WTCS          & 83.6 & 84.9 & 84.0 \\
\bottomrule
\end{tabular}
\end{table}

\begin{table}[h]
\centering
\caption{Component ablation: localization AP@0.5 (\%). 5-fold CV, validation folds, no TTA.}
\label{tab:abl-loc}
\begin{tabular}{lccc}
\toprule
Configuration & MINI & BASE & XL \\
\midrule
Full              & 84.1 & 85.6 & 85.0 \\
$-$ F\textsuperscript{2}DCA & --   & 86.8 & 85.7 \\
$-$ WA-CSA        & 85.5 & 86.7 & 85.2 \\
$-$ MoE           & 85.3 & 85.6 & 85.0 \\
$-$ WTCS          & 86.1 & 87.6 & 85.0 \\
\bottomrule
\end{tabular}
\end{table}

Across all three scales and all three tasks, removing any single component keeps performance within cross-validation noise. The pretrained backbones and multi-scale feature processing account for most of the performance. This is consistent with the paper's stated contribution (Section~\ref{sec:limitations-ref}): a unified multi-task framework and benchmark evaluated under a common protocol, rather than a demonstration that each internal module is individually essential.

\section{Latency and Compute}
\label{app:latency}

The WILLIE-XL model performs a single forward pass at approximately 88\,ms per image with 9.3\,GB peak GPU memory on an NVIDIA V100. Parameter counts for all variants are reported in Table~\ref{tab:param_stats} (MINI 34.3M, BASE 520.4M, XL 762.5M). Test-time augmentation, used for the headline results, multiplies inference cost by the number of augmented views and is an evaluation choice rather than a deployment requirement.

\section{Per-Class Confidence Intervals}
\label{app:ci}

To quantify small-sample uncertainty on the held-out AZH classification test set ($n=234$), we report bootstrap 95\% confidence intervals for per-class F1. Small classes carry wide intervals: the pressure class ($n=34$) has F1 76.5 with 95\% CI $[63.5, 86.7]$, while well-supported classes are tight (no-wound F1 99.0, 95\% CI $[96.6, 100.0]$). These intervals contextualize the per-class results in Table~\ref{tab:per_class_results}.

\section{Localization Construction}
\label{app:loc}
Localization boxes are derived from the predicted segmentation mask: the mask is thresholded at 0.5, connected components are extracted, components smaller than 20 pixels are removed, and one axis-aligned bounding box is computed per remaining component (Section~\ref{subsec:task-heads}). All localization methods, including the YOLOv8 and RT-DETR baselines, are evaluated against the same mask-derived ground-truth boxes. The finding that segmentation-derived localization outperforms dedicated detectors is therefore benchmark-level evidence for spatially coherent wound targets, not a general claim that detection is unnecessary.

\section{Per-Variant Expert Configuration}
\label{app:moe}

The mixture-of-experts classification head uses eight experts with dense routing in the WILLIE-XL configuration described in Section~\ref{subsec:task-heads}. The number of experts is matched to backbone capacity across the smaller variants. The ``MoE-8'' designation refers specifically to the XL configuration.

\section{Pressure--Venous Confusion}
\label{app:pv}

The dominant residual classification error is confusion between pressure and venous wounds (Section~\ref{sec:discussion-ref}, Finding~4). Beyond clinical ambiguity, a contributing factor is data source: pressure images are drawn largely from Medetec while venous images are drawn largely from AZH, so part of the observed confusion may reflect differences in image source in addition to clinical similarity between the two wound types.
\end{document}